Saliency detection with moving camera via background model completion

Yupei Zhang[1], Kwok-Leung Chan[2,*]
[1] Department of Electrical Engineering, City University of Hong Kong; ypzhang5-c@my.cityu.edu.hk
[2] Department of Electrical Engineering, City University of Hong Kong; itklchan@cityu.edu.hk
[*] Correspondence: itklchan@cityu.edu.hk

**Abstract**
To detect saliency in video is a fundamental step in many computer vision systems. Saliency is the significant target(s) in the video. The object of interest is further analyzed for high-level applications. The segregation of saliency and the background can be made if they exhibit different visual cues. Therefore, saliency detection is often formulated as background subtraction. However, saliency detection is challenging. For instance, dynamic background can result in false positive errors. In another scenario, camouflage will lead to false negative errors. With moving camera, the captured scenes are even more complicated to handle. We propose a new framework, called saliency detection via background model completion (SD-BMC), that comprises of a background modeler and the deep learning background/foreground segmentation network. The background modeler generates an initial clean background image from a short image sequence. Based on the idea of video completion, a good background frame can be synthesized with the co-existence of changing background and moving objects. We adopt the background/foreground segmenter, although pre-trained with a specific video dataset, can also detect saliency in unseen videos. The background modeler can adjust the background image dynamically when the background/foreground segmenter output deteriorates during processing of a long video. To the best of our knowledge, our framework is the first one to adopt video completion for background modeling and saliency detection in videos captured by moving camera. The results, obtained from the PTZ videos, show that our proposed framework outperforms some deep learning-based background subtraction models by 11% or more. With more challenging videos, our framework also outperforms many high ranking background subtraction methods by more than 3%.

## 1. Introduction
High-level applications such as human motion analysis [1] and intelligent transportation system [2] demand the localization of targets in the video. For instance, in video surveillance, humans are detected for motion recognition. Vehicles are located in intelligent transportation system. This foremost task can be achieved via saliency detection. Assuming that the background scene possesses invariant characteristics, target is detected due to its deviated visual cues. One approach is to formulate the task as background/foreground segmentation. With the estimated background scene model, the foreground (i.e. saliency) is segmented by a pixelwise background subtraction algorithm.

However, the two assumptions – invariant background and deviation of foreground, may be violated in some circumstances. For instances, background motion and illumination change can result in false detection (false positive). With the pre-generated background model, background pixels may be predicted as foreground pixels. This is false positive error which is due to image feature(s) of the background pixel different from the background model. On the other hand, camouflage and intermittent object motion will lead to missing detection. This false negative error is due to the fact that some foreground pixels may be erroneously identified as the background pixels if they have similar image feature(s) to the background model.

The saliency detection task becomes more challenging with the use of moving camera. The assumption of a static background is violated. Videos can be captured by a pan-tilt-zoom (PTZ) camera or free-moving (e.g. hand-held) camera. Systems that can handle such type of video are of interest. They demand sophisticated techniques for generating and maintaining the background model, as well as foreground segmentation.

Researchers have proposed various background subtraction algorithms. Many background subtraction methods are deterministic, i.e. background/foreground segmentation is achieved based on hand-crafted features. One earliest approach is to adopt statistical models [3, 4]. Elgammal et al. [5] utilized kernel estimator to characterize the probability density function (pdf) of the background pixels. Some researchers have presented the survey on the background subtraction techniques [6, 7]. Sobral and Vacavant [8] evaluated 29 background subtraction methods. In general, background subtraction comprises of three main parts – background modeler, background/foreground classifier, and background updating.

Another approach is to use neural network for saliency detection. Its cognitive power is made possible with the structure simulating the complex connectivity of neurons. Maddalena and Petrosino [9] proposed the Self Organizing Background Subtraction (SOBS). The background scene is modeled with the weights of the neurons. The network compares the current image frame with the background model and outputs the pixelwise background/foreground classification. Recently, a popular approach is to develop deep learning models, such as convolutional neural network (CNN). The layered structure can accommodate the multi-scale representation, with which image data are transformed and abstract features are extracted. Wang et al. [10] proposed a basic CNN model, with which multi-resolution CNN and cascaded CNN architectures were designed for object segmentation. Lim and Keles [11], proposed an encoder-decoder network for object segmentation. The encoder part is a triple CNNs for multi-scale feature extraction. The concatenated feature map is fed to a transposed convolutional network in the decoder part. They [12] further proposed another model which uses feature pooling module on top of the encoder part.

In this paper, we propose a new framework, called saliency detection via background model completion (SD-BMC), that comprises of a background modeler and the deep learning background/foreground segmentation network. Our framework can detect saliency in videos captured by moving camera. The results, obtained from the benchmark datasets, show that our proposed framework outperforms many high-ranking background subtraction models. Figure 1 shows an overview of SD-BMC which performs two main tasks – generation of initial background model, and continuous saliency detection with updating of background model. Our contributions can be summarized as follows:

- Inspired by the filling of missing pixels via the inpainting technique, we adopt the video completion module for modeling the background scene. To generate a clean background frame, foreground objects will be substituted by the estimated background colors. Guided by the optical flow, the video completion module can generate good background model for video captured by moving camera, which is not possible for other existing methods.
- We adopt the BSUV-Net 2.0 [13] for background/foreground segmentation. Although the model is pre-trained with the CDNet [14] video dataset, it can also segment foreground in unseen videos. However, most of the videos in CDNet are captured by static camera. BSUV-Net 2.0 still produces some FP and FN errors in moving camera videos. Therefore, we replace the background frame generation method of BSUV-Net 2.0 with our video completion based background modeler.
- We propose a framework that comprises of the video completion-based background modeler and the enhanced BSUV-Net 2.0 foreground segmentation network. To thoroughly evaluate the new framework, we create our own video dataset with videos captured by PTZ camera and free-moving camera. The results show that our framework outperforms many high-ranking background subtraction models.

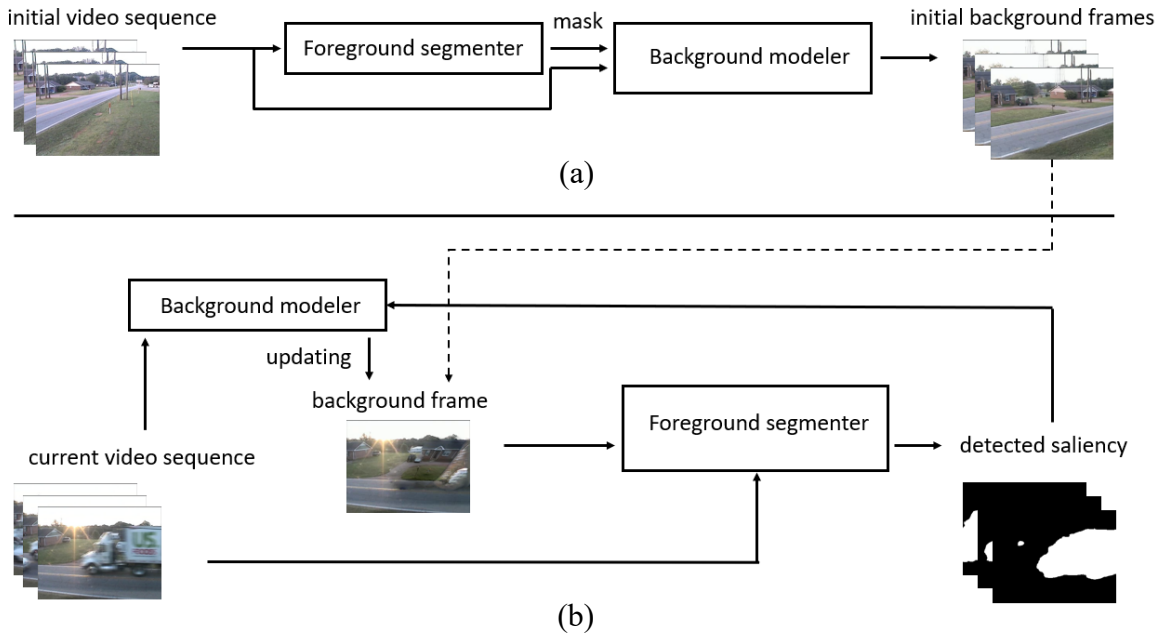

Figure 1. Overview of the saliency detection framework: (a) background model initialization; (b) continuous saliency detection.

The paper is organized as follows. The related researches on background subtraction, in particular with videos captured by moving camera, are reviewed in the following section. Section 3 elaborates our saliency detection framework. We compare our framework with other high-ranking background subtraction algorithms. Quantitative and visual results are presented in section 4. Discussion is also made on the performance of all these methods. Finally, we draw the conclusion in section 5.

## 2. Related work

Many methods have been proposed for segmenting foreground in videos captured by stationary cameras. In this section, we review sophisticated methods that are proposed to handle videos captured by moving cameras. Moving cameras can be categorized into two types: constrained moving camera, and freely moving camera. For instance, PTZ camera belongs to the first category. In the second category, examples are hand-held camera, smartphone, or camera mounted on drone. Methods developed for constrained camera may not perform well with freely moving camera.

Hishinuma et al. [15] considered the camera small pan/tilt motion as translational. The translation amount is computed from the correlation of the FFT phase terms of stationary background blocks. The synthesized still background model is then used for foreground segmentation. In [16], camera motion is compensated by calculating the homography transformation between two image frames. Scene model, which is a panoramic background, is then generated from the motion compensated video. Foreground objects are detected by comparing the panoramic background with individual image frames of the video. Szolgay et al. [17] proposed a method for detecting moving objects in video taken by a wearable camera. Global camera motion is estimated first by a hierarchical block matching algorithm and then refined by a robust motion estimator. Foreground is identified as the difference between motion-compensated image frames. Tao and Ling [18] proposed a neural network for segmenting foreground in videos captured by PTZ cameras. Deep learning features are extracted by a pre-trained network. Homography matrix is estimated from previous image frames and current image frame with a semantic attention based deep homography estimator. The warped previous frames, current frames, and their features are fed into the fusion network for foreground mask prediction. Komagal and Yogameena [19] reviewed the methods and the datasets for foreground segmentation research with PTZ camera.

With the use of freely-moving camera, both background and foreground are changing. The assumption of background modeling may be violated. For instance, when background and foreground motions are similar, the background model is contaminated with foreground colors. In another scenario, inaccurate camera motion estimation will give rise to false positive errors. Yun et al. [20] proposed an adaptive scheme that can update the background model in accordance with the changes of background. The scheme compensates three types of change – background motion produced by moving camera, foreground motion, and illumination change. Knowing that an explicit camera motion model is not reliable, Sajid et al. [21] proposed an online framework such that both background and foreground models are continuously updated. Background motion is estimated with a low-rank approximation. Motion and appearance models are combined to produce the background/foreground classification. Zhu and Elgammal [22] proposed a multi-layered framework for background subtraction. In each layer, both motion and appearance model are estimated and used for foreground detection. Probability map is inferred by kernel density estimator [5]. Finally, segmented foreground is generated from the multi-layered outputs by multi-label graph-cut. Chapel and Bouwmans [23] reviewed the moving object detection methods with moving camera. They grouped the methods into two categories in accordance with scene representation – single-plane and multi-plane. Methods in the first group may generate a panoramic background by image mosaic. Some methods detect moving objects via motion segmentation. Multi-plane approach estimates several planes (may be real or not) as scene representation. Matched feature points are located and eventually used for background/foreground classification.

We adopt the background-centric approach for saliency detection. Instead of modeling the background based on camera motion compensation which may be inaccurate, we generate and update the background dynamically via video completion and continuous monitoring of foreground segmentation result. Our background modeler, based on the optical flow information, can generate a much better background frame for video captured by a moving camera than other methods. As demonstrate in our results, our framework with the cascade of background modeler and deep-learning foreground segmenter, outperforms many high-ranking background subtraction models in saliency detection.

Various video datasets were created for background subtraction research. The CDNet 2014 dataset [14] contains videos grouped under 11 categories. Each video record provides the original image sequence and the corresponding ground truths. Many of the videos were captured in different challenging scenes. For instance, the "PTZ" category contains four videos captured by PTZ camera. The Hopkins 155 dataset [24] contains indoor and outdoor panning videos. Perazzi et al. [25] proposed three versions of the Densely Annotated Video Segmentation (DAVIS) dataset. Some videos were captured by shaking camera. The SegTrack v2 dataset [26] contains videos captured by moving camera with ground truth of moving object. Labeled and Annotated Sequences for Integral Evaluation of SegmenTation Algorithms (LASIESTA) [27] contains real indoor and outdoor videos with pan, tilt or shaking cameras. In our experimentations, we create our own dataset which comprises of videos captured by PTZ camera and moving camera extracted from various publicly available video datasets.

## 3. Saliency detection framework

The saliency detection framework SD-BMC is shown in Figure 1. First, we initialize the system with the first 100 frames. We use the background image generated by the foreground segmenter (BSUV-Net 2.0 [13]) with median filter to create masks in this step. These masks, together with the initial image sequence, will be put into the video completion-based background modeler (FGVC [28]). From the sequence of completed frames, the most recent one is selected as the background. In the stream of saliency detection, the initial background frame and the current image sequence will be input to the foreground segmenter. The background model will be updated based on the foreground segmentation result. The stream of saliency detection with feedback will continue until all the video frames are processed.

### 3.1 Background modeler

Many background modeling algorithms can estimate a clean background frame, even the image sequence contains moving objects. However, if the foreground objects exist too long, there will be phenomena like ghosts in the background image. The problem becomes more complicated with video captured by a moving camera. Deep learning-based methods have been proposed for background modeling. For instance, Farnoosh et al. [29] proposed a variational autoencoder (VAE) framework for background estimation from videos recorded by fixed camera. In our experimentation on videos of moving camera, there are always blur pixels existing in the final background images.

We adopt and modify the video completion method FGVC [28] for background modeling. The algorithm can generate a clean background image with more attention to the masks corresponding to the foreground objects and also the changing scene between adjacent image frames. Figure 2 shows our video completion-based background modeler. In part (a), the color video sequence and the corresponding binary masks are input to the background modeler. The masks are the foreground regions that need to be completed. Next, in part (b), optical flow between adjacent frames is computed with FlowNet2 [30]. Moreover, flow between some non-adjacent frames is also computed. This can help to estimate the missing background colors when camera motion is large. The background flow is predicted from the color video sequence, while the foreground flow is predicted from the masks. In each flow field, flow edges are extracted. Guided by the flow edge map, a completed optical flow field is generated. In part (c), a set of candidate pixels are computed for each missing pixel. Most of the missing pixels can be filled with inpainting via fusion of the candidate pixels. After that, the network will use Poisson reconstruction to generate the initial completed background frame. Finally, in the last part (d), the modeler will fix the remaining missing pixels with a number of inpainting iterations until there is no missing pixel.

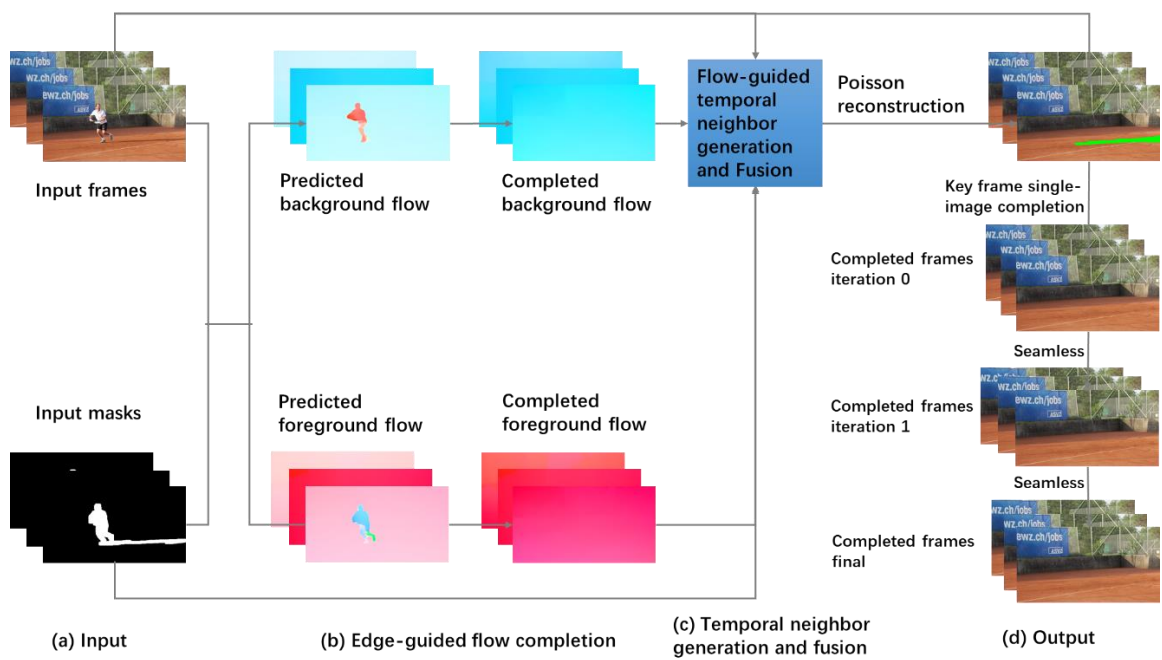

Figure 2. Video completion-based background modeler.

Experimentation is performed to determine the length of the video sequence for background generation. If it is too short, there may not be enough number of candidate pixels for background color synthesis. If it is too long, the time for background generation will be long, and the actual saliency detection will be delayed. First, we choose 30 frames for background generation. Figure 3 shows the background modeling on one video. It can be seen that ghosts exist in the background frame. Then, we lengthen the initialization sequence to 100 frames. Most of the foreground pixels can be substituted with the background colors. Table 1 compares the F-measure of saliency detection on the PTZ category of CDNet 2014 dataset. Finally, we fix the length to 100 frames.

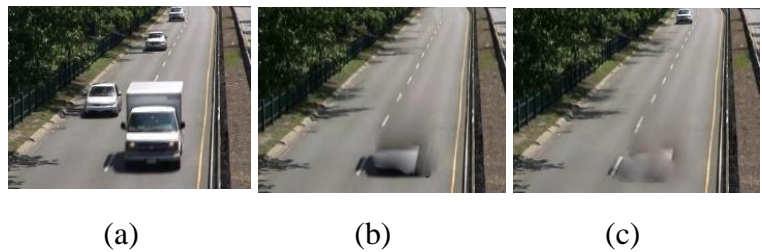

(a)  (b)  (c)

Figure 3. Visual results of background modeling: (a) original frame; (b) 30 initialization frames; (c) 100 initialization frames.

Table 1. F-measure of saliency detection with two settings for background modeling.

| Length of initialization sequence | F-measure |
|---|---|
| 30 frames | 0.8062 |
| 100 frames | **0.8147** |

3.2 Foreground segmentation

We adopt BSUV-Net 2.0 [13] as foreground segmenter. As shown in Figure 4, it has a U-Net [31] like structure. Based on BSUV-Net [32], BSUV-Net 2.0 further improves the background subtraction performance on complicated videos with more spatio-temporal data augmentations. The encoder-decoder structure contains five convolutional blocks in the downsampling path, four convolutional blocks in the upsampling path, and their links via concatenation. The detail of the configuration is shown in Table 2.

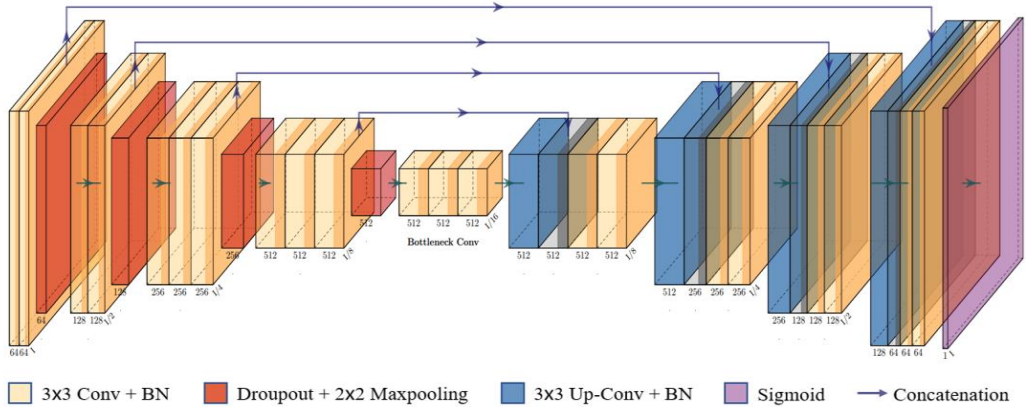

Figure 4. Structure of foreground segmenter.

Table 2. Layer configuration of foreground segmenter (SD: spatial dropout layer; BN: batch normalization).

| Network structure of BSUV | kernel_size | in_channels | out_channels |
| --- | --- | --- | --- |
| Conv + BN | 3 x 3 | 12 | 64 |
| Conv + BN | 3 x 3 | 64 | 64 |
| SD + Maxpooling | 2 x 2 | 64 | 64 |
| Conv + BN | 3 x 3 | 64 | 128 |
| Conv + BN | 3 x 3 | 128 | 128 |
| SD + Maxpooling | 2 x 2 | 128 | 128 |
| Conv + BN | 3 x 3 | 128 | 256 |
| Conv + BN | 3 x 3 | 256 | 256 |
| Conv + BN | 3 x 3 | 256 | 256 |
| SD + Maxpooling | 2 x 2 | 256 | 256 |
| Conv + BN | 3 x 3 | 256 | 512 |
| Conv + BN | 3 x 3 | 512 | 512 |
| Conv + BN | 3 x 3 | 512 | 512 |
| SD + Maxpooling | 2 x 2 | 512 | 512 |
| Conv + BN | 3 x 3 | 512 | 512 |
| Conv + BN | 3 x 3 | 512 | 512 |
| Conv + BN | 3 x 3 | 512 | 512 |
| Up-conv + BN | 3 x 3 | 512 | 512 |
| Conv + BN | 3 x 3 | 512+512 | 512 |
| Conv + BN | 3 x 3 | 512 | 512 |
| Up-conv + BN | 3 x 3 | 512 | 512 |
| Conv + BN | 3 x 3 | 512+256 | 256 |
| Conv + BN | 3 x 3 | 256 | 256 |
| Up-conv + BN | 3 x 3 | 256 | 256 |
| Conv + BN | 3 x 3 | 256+128 | 128 |
| Conv + BN | 3 x 3 | 128 | 128 |
| Up-conv + BN | 3 x 3 | 128 | 128 |
| Conv + BN | 3 x 3 | 128+64 | 64 |
| Conv + BN | 3 x 3 | 64 | 64 |
| Conv + BN | 3 x 3 | 64 | 1 |
| Sigmoid | | 1 | 1 |

An empty background frame, a recent background frame, the current frame and corresponding foreground probability maps (FPM) are needed for background/foreground separation. The input has a total of 12 channels. To avoid overfitting problems and increase the generalization of the network, it uses a batch normalization layer for each convolution layer in the encoder part and a convolution transpose layer in the decoder part. Dropout layers are also used before max-pooling to make the network more generative. Finally, the network uses the sigmoid function to get the prediction value of the pixels in the output binary saliency detection.

Tezcan et al. [13] simulated some changes, e.g. changes that look like videos captured by PTZ camera, for data augmentation in training the model. However, as shown in our experimental results, BSUV-Net 2.0 is still not good enough in saliency detection with videos captured by PTZ camera and freely-moving camera. It is because the background modeling method cannot generate a fairly good background frame for complicated videos. Therefore, in our saliency detection stream, we disable the default background modeling method. Instead, we use the video completion-based background modeler which can generate a better background frame.

## 4. Result and discussion

*4.1 Datasets*

We test our saliency detection framework on CDNet 2014 dataset [14] and our customized dataset. As shown in Table 3, CDNet 2014 comprises of 11 categories, each of which contains 4 to 6 videos. Each video record provides the original image sequence and the corresponding ground truths. Some videos, e.g. the PTZ category, were captured in challenging scenes.

Table 3. CDNet 2014 dataset categories and video scene.

| Category | Video scenes | | | | | |
|---|---|---|---|---|---|---|
| Bad Weather | blizzard (7000 frames) | skating (3900 frames) | snowFall (6500 frames) | wetSnow (3500 frames) | | |
| Low Framerate | port 0 17fps (3000 frames) | tramCrossroad 1fps (900 frames) | tunnelExit 0 35fps (4000 frames) | turnpike 0 5fps (1500 frames) | | |
| Night Videos | bridgeEntry (2500 frames) | busyBoulvard (2760 frames) | fluidHighway (1364 frames) | streetCornerAtNight (5200 frames) | tramStation (3000 frames) | winterStreet (1785 frames) |
| PTZ | continuousPan (1700 frames) | intermittentPan (3500 frames) | twoPositionPTZCam (2300 frames) | zoonInZoomOut (1130 frames) | | |
| Thermal | corridor (5400 frames) | library (4900 frames) | park (600 frames) | diningRoom (3700 frames) | lakeSide (6500 frames) | |
| Shadow | backdoor (2000 frames) | bungalows (1700 frames) | busStation (1250 frames) | cubicle (7400 frames) | peopleInShade (1199 frames) | copyMachine (3400 frames) |
| Intermittent Object Motion | abandonedBox (4500 frames) | parking (2500 frames) | streetLight (3200 frames) | sofa (2750 frames) | tramstop (3200 frames) | winterDriveway (2500 frames) |
| Camera Jitter | badminton (1150 frames) | boulevard (2500 frames) | sidewalk (1200 frames) | traffic (1570 frames) | | |
| Dynamic Background | boats (7999 frames) | canoe (1189 frames) | fountain01 (1184 frames) | fountain02 (1499 frames) | overpass (3000 frames) | fall (4000 frames) |
| Baseline | highway (1700 frames) | office (2050 frames) | pedestrians (1099 frames) | PETS2006 (1200 frames) | | |
| Turbulence | turbulence0 (5000 frames) | turbulence1 (4000 frames) | turbulence2 (4500 frames) | turbulence3 (2200 frames) | | |

Our customized dataset comprises of 22 videos from the FBMS dataset [33] and 8 videos from the LASIESTA dataset [27]. The videos were captured by handheld cameras and PTZ cameras. For videos selected from the FBMS dataset, we used manually define the ground truths for 20 continuous frames randomly chosen after the 100th frame in each video. For videos from the LASIESTA dataset, each record provides a number of ground truth images. Table 4 shows the details of our customized dataset.

Table 4. Customized dataset categories and video scene.

| category\method | video | number of frames | ground truth frames |
|---|---|---|---|
| animals | cats06 | 331 | (190, 209) |
|  | cats07 | 193 | (110, 129) |
|  | dogs02 | 420 | (160, 179) |
|  | horses01 | 500 | (280, 299) |
|  | horses02 | 240 | (140, 159) |
|  | horses03 | 240 | (214, 233) |
|  | horses04 | 800 | (540, 559) |
|  | horses05 | 456 | (320, 339) |
|  | horses06 | 720 | (380, 399) |
|  | rabbits01 | 290 | (200, 219) |
| people | I_MC_01 | 300 | (1, 300) |
|  | I_SM_01 | 300 | (1, 300) |
|  | I_SM_02 | 300 | (1, 300) |
|  | I_SM_03 | 300 | (1, 300) |
|  | marple1 | 328 | (309, 328) |
|  | marple2 | 250 | (220, 239) |
|  | marple3 | 323 | (250, 269) |
|  | marple6 | 800 | (180, 199) |
|  | marple7 | 528 | (370, 389) |
|  | marple10 | 460 | (440, 459) |
|  | marple11 | 173 | (150, 169) |
|  | O_MC_01 | 425 | (1, 425) |
|  | O_SM_01 | 425 | (1, 425) |
|  | O_SM_02 | 425 | (1, 425) |
|  | O_SM_03 | 425 | (1, 425) |
|  | people03 | 180 | (160, 179) |
|  | people04 | 320 | (270, 289) |
|  | people05 | 260 | (140, 159) |
| things | farm01 | 252 | (194, 213) |
|  | tennis | 466 | (281, 300) |

*4.2 Evaluation metrics*

To evaluate the performance of our framework and other baseline methods, we compute 7 quantitative measures: *Recall, Specificity, False Positive Rate (FPR), False Negative Rate (FNR),* Percentage of Wrong Classifications (*PWC*), *F-Measure*, and *Precision* (where *TP* is true positive, *FP* is false positive, *FN* is false negative, and *TN* is true negative).

$$\mathrm{Recall} = \frac{TP}{TP + FN} \qquad (1)$$

$$Specificity = \frac{TN}{TN + FP} \qquad (2)$$

$$FPR = \frac{FP}{FP + TN} \qquad (3)$$

$$FNR = \frac{FN}{TP + FN} \qquad (4)$$

$$PWC = \frac{100 \times (FN + FP)}{TP + FN + FP + TN} \tag{5}$$

$$F - Measure = \frac{2 \times \text{Precision} \times \text{Recall}}{\text{Precision} + \text{Recall}} \tag{6}$$

$$\text{Precision} = \frac{TP}{TP + FP} \tag{7}$$

*4.3 Performance evaluation*
We implement SD-BMC with the Python-based Pytorch. The computing platform comprised of Intel Xeon Silver 4108 CPU 1.8G 16 Cores, and a HPC Cluster with NVIDIA RTX 2080Ti 11GB x 8 GPU nodes. The background frame, either in the initialization or in the updating process, is generated from a sequence of 100 image frames. Therefore, each video is partitioned into sections of 100 frames. If the last section contains less than 100 frames, we input all the remaining frames into our framework. We resize the original image sequence and the ground truth images with the resolution of 320 * 240.

We compare SD-BMC with six background subtraction algorithms – BSUV-Net [32], BSUV-Net 2.0 [13], Fast BSUV-Net 2.0 [13], PAWCS [34], SuBSENSE [35], and ViBe [36]. Tezcan et al. [32] first proposed the BSUV-Net. Background frames are estimated from the video. The current frame of the video and the background frames are input to the fully-convolutional neural network for background subtraction. They proposed the second version of the model [13] by training with data simulating spatio-temporal changes. Moreover, they developed the Fast BSUV-Net 2.0 [13] which is a real-time version of the model. St-Charles et al. proposed SuBSENSE [35] and PAWCS [34] for change detection. The background model is a codebook which is generated based on the persistence of pixel features. They are among the high-ranking methods in CDNet 2014. Barnich et al. [36] adopted the bag of words approach and proposed an efficient background subtraction method ViBe. At each pixel location, some samples are randomly selected from the image sequence and stored as background colors. The background model is also updated with a random process.

*4.4 Quantitative and visual results*
Table 5 shows the numerical results of SD-BMC on CDNet 2014 dataset. Table 6 shows the average results of BSUV-Net, BSUV-Net 2.0, Fast BSUV-Net 2.0, and SD-BMC. The bold numbers represent the best results. Table 7 compares the results of BSUV-Net, BSUV-Net 2.0, Fast BSUV-Net 2.0, and SD-BMC on the PTZ category of CDNet 2014. Figure 5 shows some visual results of BSUV-Net 2.0 and SD-BMC.

Table 5. Evaluation metrics of SD-BMC on CDNet 2014.

| Category \ Metric | Recall | Specificity | FPR | FNR | PWC | F-Measure | Precision |
|---|---|---|---|---|---|---|---|
| PTZ | 0.8798 | 0.9970 | 0.0030 | 0.1202 | 0.3788 | 0.8147 | 0.7880 |
| badWeather | 0.7139 | 0.9997 | 0.0003 | 0.2861 | 0.4538 | 0.8155 | 0.9790 |
| baseline | 0.9325 | 0.9987 | 0.0013 | 0.0675 | 0.2642 | 0.9514 | 0.9736 |
| cameraJitter | 0.8310 | 0.9973 | 0.0027 | 0.1690 | 0.9369 | 0.8790 | 0.9382 |
| dynamic background | 0.7437 | 0.9998 | 0.0002 | 0.2563 | 0.1229 | 0.8013 | 0.9336 |
| intermittentObjectMotion | 0.7987 | 0.9980 | 0.0020 | 0.2013 | 1.3093 | 0.8758 | 0.9809 |
| lowFramerate | 0.5614 | 0.9994 | 0.0006 | 0.4386 | 0.6950 | 0.6292 | 0.9046 |
| nightVideos | 0.4856 | 0.9992 | 0.0008 | 0.5144 | 1.1114 | 0.5727 | 0.9328 |
| shadow | 0.9389 | 0.9984 | 0.0016 | 0.0611 | 0.4378 | 0.9543 | 0.9704 |
| thermal | 0.6156 | 0.9990 | 0.0010 | 0.3844 | 1.2019 | 0.7075 | 0.9531 |
| turbulence | 0.5732 | 0.9998 | 0.0002 | 0.4268 | 0.2583 | 0.7107 | 0.9654 |
| Average value | 0.7340 | 0.9988 | 0.0012 | 0.2660 | 0.6518 | 0.7920 | 0.9381 |

Table 6. Average evaluation metrics of BSUV-Net, BSUV-Net 2.0, Fast BSUV-Net 2.0, and SD-BMC on CDNet 2014.

| Method | Recall | Specificity | FPR | FNR | PWC | F-Measure | Precision |
|---|---|---|---|---|---|---|---|
| BSUV-Net | **0.8203** | 0.9946 | 0.0054 | **0.1797** | 1.1402 | 0.7868 | 0.8113 |
| BSUV-Net 2.0 | 0.8136 | 0.9979 | 0.0021 | 0.1864 | 0.7614 | **0.8387** | 0.9011 |
| Fast BSUV-Net 2.0 | 0.8181 | 0.9956 | 0.0044 | 0.1819 | 0.9054 | 0.8039 | 0.8425 |
| SD-BMC | 0.7340 | **0.9988** | **0.0012** | 0.2660 | **0.6518** | 0.7920 | **0.9381** |

Table 7. Evaluation metrics of BSUV-Net, BSUV-Net 2.0, Fast BSUV-Net 2.0, and SD-BMC on PTZ category of CDNet 2014.

| Method | Recall | Specificity | FPR | FNR | PWC | F-Measure | Precision |
|---|---|---|---|---|---|---|---|
| BSUV-Net | 0.8045 | 0.9909 | 0.0091 | 0.1955 | 1.0716 | 0.6282 | 0.5897 |
| BSUV-Net 2.0 | 0.7932 | 0.9957 | 0.0043 | 0.2068 | 0.5892 | 0.7037 | 0.6829 |
| Fast BSUV-Net 2.0 | 0.8056 | 0.9878 | 0.0122 | 0.1944 | 1.3516 | 0.5014 | 0.4236 |
| SD-BMC | **0.8798** | **0.9970** | **0.0030** | **0.1202** | **0.3788** | **0.8147** | **0.7880** |

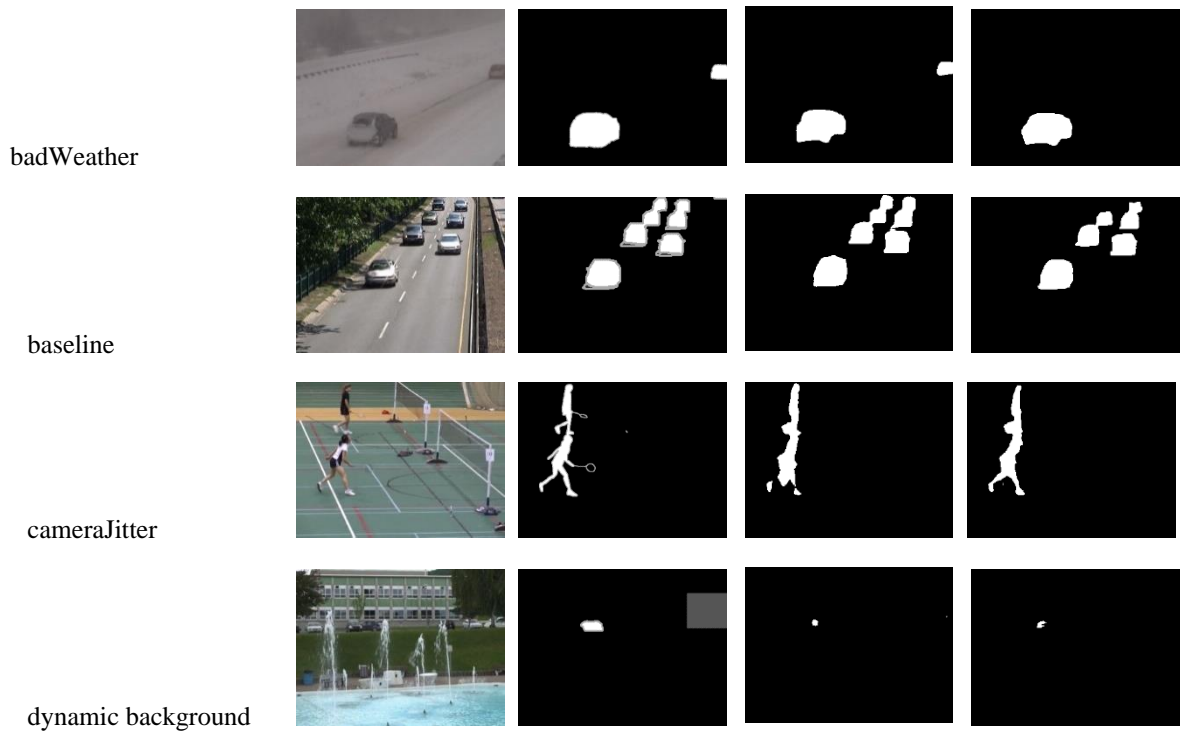

badWeather

baseline

cameraJitter

dynamic background

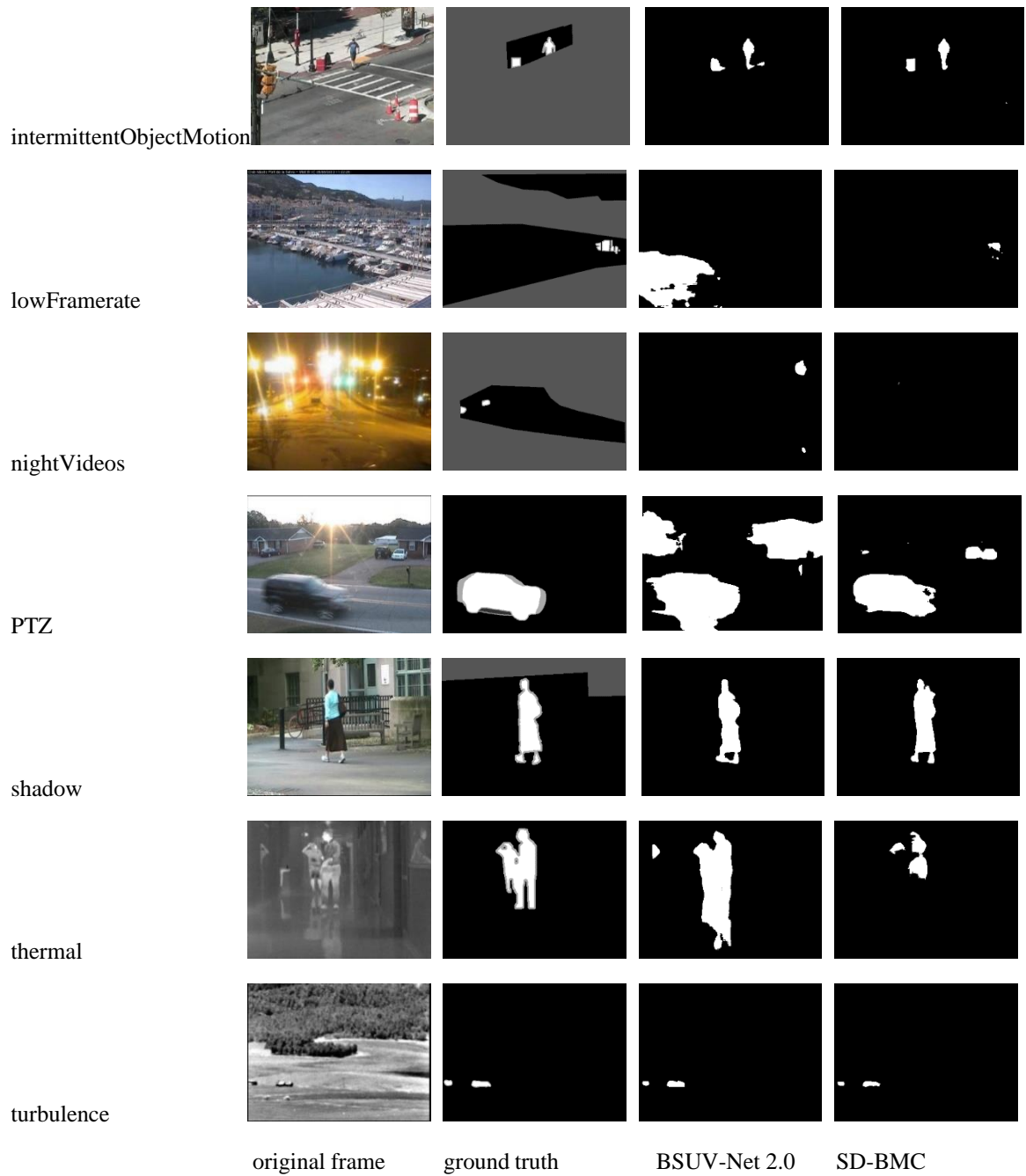

Figure 5. Visual results of BSUV-Net 2.0 and SD-BMC on CDNet 2014.

Our saliency detection framework achieves the best average *Recall, FPR, PWC*, and *Precision* on CDNet 2014 dataset. As shown in the visual results, SD-BMC can achieve comparable performance as BSUV-Net 2.0 in many video categories. The superiority of SD-BMC over BSUV-Net 2.0 can be seen in the PTZ category. Due to the poor background frame, BSUV-Net 2.0 produces large amount of *FP* errors. The quantitative results shown in Table 7 clearly indicate that SD-BMC outperforms the other three models in all evaluation metrics on PTZ videos. When a single numeric result, *F-measure*, is chosen for ranking, SD-BMC outperforms all other methods by more than 11%.

Table 8 shows the average results of BSUV-Net, BSUV-Net 2.0, PAWCS, SuBSENSE, ViBe, and SD-BMC on customized dataset. Table 9 compares the *F-measure* on individual videos. Figure 6 shows some visual results of BSUV-Net, BSUV-Net 2.0, PAWCS, SuBSENSE, ViBe, and SD-BMC.

Table 8. Average evaluation metrics of BSUV-Net, BSUV-Net 2.0, PAWCS, SuBSENSE, ViBe, and SD-BMC on customized dataset.

| Method | Recall | Specificity | FPR | FNR | PWC | F-Measure | Precision |
|---|---|---|---|---|---|---|---|
| BSUV-Net | 0.4823 | 0.8556 | 0.1444 | 0.5177 | 18.4314 | 0.3707 | 0.4547 |
| BSUV-Net 2.0 | 0.5273 | **0.9934** | **0.0066** | 0.4727 | 7.3303 | 0.6123 | **0.8827** |
| PAWCS | 0.4730 | 0.8695 | 0.1305 | 0.5270 | 19.0191 | 0.3323 | 0.3055 |
| SuBSENSE | 0.3066 | 0.9460 | 0.0540 | 0.6934 | 13.9365 | 0.2936 | 0.4167 |
| ViBe | 0.1332 | 0.7947 | 0.2053 | 0.6680 | 25.4062 | 0.2001 | 0.1707 |
| SD-BMC | **0.5635** | 0.9906 | 0.0094 | **0.4365** | **7.1580** | **0.6446** | 0.8719 |

Table 9. *F-measure* of BSUV-Net, BSUV-Net 2.0, PAWCS, SuBSENSE, ViBe, and SD-BMC on individual videos.

| F-Measure comparison on individual video | Video | SD-BMC | BSUV-Net 2.0 | BSUV-Net | PAWCS | SuBSENSE | ViBe |
|---|---|---|---|---|---|---|---|
| animals | cats06 | **0.4417** | 0.0041 | | | 0.0005 | 0.0062 |
| | cats07 | 0.5700 | 0.6634 | 0.5549 | **0.7506** | 0.4268 | 0.4408 |
| | dogs02 | **0.7758** | 0.6531 | 0.1754 | 0.4514 | 0.6144 | 0.0981 |
| | horses01 | 0.7163 | 0.7075 | 0.5733 | 0.5131 | **0.7398** | 0.2828 |
| | horses02 | 0.3591 | **0.4368** | 0.3474 | 0.1090 | 0.0264 | 0.0485 |
| | horses03 | **0.8341** | 0.8130 | 0.1235 | 0.1844 | 0.2957 | 0.0454 |
| | horses04 | 0.2322 | 0.2198 | **0.4269** | 0.0018 | 0.2025 | 0.0449 |
| | horses05 | 0.1433 | 0.1493 | 0.6209 | 0.4218 | 0.1793 | **0.6971** |
| | horses06 | **0.4626** | 0.4583 | 0.1986 | 0.2434 | 0.4240 | 0.0353 |
| | rabbits01 | **0.1972** | 0.1607 | 0.1805 | 0.2484 | 0.1214 | 0.0239 |
| | **Average** | **0.4732** | 0.4266 | 0.3557 | 0.3249 | 0.3031 | 0.1723 |
| people | I_MC_01 | 0.8343 | **0.8495** | 0.6551 | 0.3541 | 0.4782 | 0.1049 |
| | I_SM_01 | 0.7957 | **0.8631** | 0.2943 | 0.4522 | 0.6024 | 0.2066 |
| | I_SM_02 | 0.7577 | **0.8310** | 0.3354 | 0.4601 | 0.5193 | 0.2312 |
| | I_SM_03 | 0.7283 | **0.8248** | 0.3593 | 0.4367 | 0.4462 | 0.2334 |
| | marple1 | **0.6106** | 0.6042 | 0.3784 | 0.2276 | 0.3022 | 0.0946 |
| | marple2 | 0.8292 | **0.9490** | 0.0586 | 0.0312 | 0.0085 | 0.4303 |
| | marple3 | **0.9186** | 0.4303 | 0.0013 | 0.0661 | 0.0050 | 0.2659 |
| | marple6 | **0.4521** | 0.3793 | 0.2967 | 0.3744 | 0.4677 | 0.3463 |
| | marple7 | **0.9308** | 0.8645 | 0.5030 | 0.0032 | 0.0452 | 0.0895 |
| | marple10 | 0.6870 | **0.8286** | 0.0016 | 0.0429 | 0.0443 | 0.0540 |
| | marple11 | 0.4359 | **0.7044** | 0.4371 | 0.1585 | 0.0344 | 0.3260 |
| | O_MC_01 | **0.7097** | 0.6292 | 0.2610 | 0.3230 | 0.2195 | 0.1570 |
| | O_SM_01 | 0.8686 | **0.8704** | 0.3575 | 0.3559 | 0.2926 | 0.1446 |
| | O_SM_02 | 0.8113 | **0.8298** | 0.1920 | 0.3108 | 0.3205 | 0.1067 |
| | O_SM_03 | 0.7955 | **0.8115** | 0.1516 | 0.2569 | 0.2894 | 0.0909 |
| | people03 | **0.5481** | 0.4985 | 0.2137 | 0.4588 | 0.0144 | 0.2158 |
| | people04 | 0.8333 | **0.8457** | 0.3818 | 0.1923 | 0.2498 | 0.0628 |
| | people05 | **0.8786** | 0.8662 | 0.6771 | 0.7264 | 0.6623 | 0.3420 |
| | **Average** | 0.7459 | **0.7489** | 0.3086 | 0.2906 | 0.2779 | 0.1946 |
| things | farm01 | **0.5936** | 0.4329 | 0.1859 | 0.3668 | 0.1654 | 0.2630 |
| | tennis | 0.8360 | **0.8902** | 0.6442 | 0.2836 | 0.4341 | 0.2035 |
| | **Average** | **0.7148** | 0.6616 | 0.4151 | 0.3252 | 0.2998 | 0.2333 |
| **Total Average** | | **0.6446** | 0.6123 | 0.3707 | 0.3323 | 0.2936 | 0.2001 |

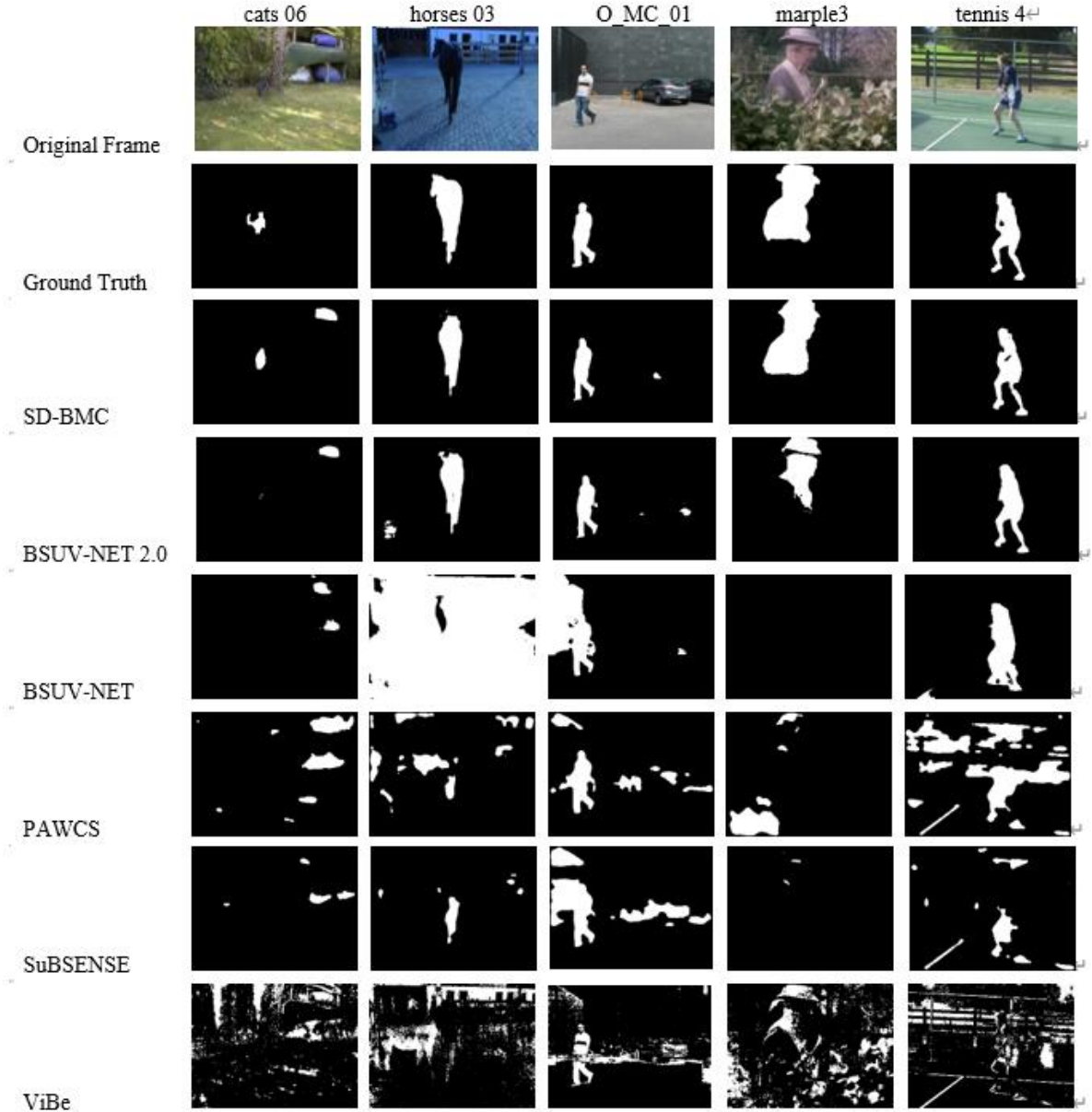

Figure 6. Visual results on customized dataset.

The videos in the customized dataset are more challenging. We classify the videos, in according to their contents, into 3 groups: animals, people, and things. SD-BMC achieves the best average *Recall, FNR, PWC*, and *F-Measure*. We select a single numeric result, *F-measure*, for assessing the performance of all methods on individual videos. As shown in Table 9, SD-BMC achieves the best *F-measure* in many videos. The average *F-measure* in "animals" and "things" groups are higher than all other methods, while in "people" group the average *F-measure* is slightly lower than BSUV-Net 2.0. As shown in Figure 6, SD-BMC can detect saliency very close to the ground truth. The second best method, BSUV-Net 2.0, produces more *FP* and *FN* errors. Overall, SD-BMC outperforms BSUV-Net 2.0 by more than 3%.

*4.5 Comparative analysis*

According to the results on the CDNet 2014 dataset, we found that SD-BMC outperforms other methods in the PTZ category. For other video categories, SD-BMC achieves comparable results with other methods. The reason is that our video completion-based background modeler, together with the feedback scheme, can generate clear and updated background images. FGVC is a non-scene-specific method, which could generalize to unseen videos. As FGVC can capture temporal and spatial information, this modeler can generate much better background images. On contrary, the empty background images used in BSUV-Net 2.0 are very blur, which could significantly affect the final saliency detection result. PAWCS and SuBSENSE, which are designed for fixed camera, produce even worse background frame. Figure 7 shows the comparison of the background images.

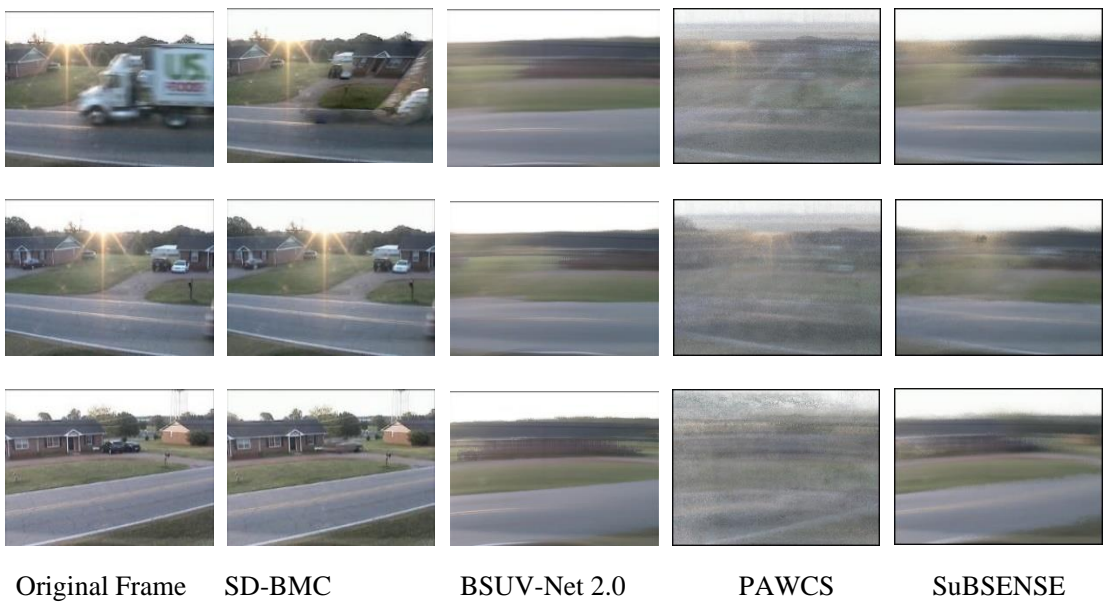

   Original Frame    SD-BMC        BSUV-Net 2.0        PAWCS        SuBSENSE

Figure 7. Comparison of background frames used in BSUV-Net 2.0, PAWCS, SuBSENSE, and SD-BMC on PTZ video.

In Table 9, the *F-Measure* values of BSUV-Net and PAWCS on "cats06" video are left blank. It is because these two methods have zero *Recall* value. Therefore, their *F-Measure* cannot be calculated. It is clear that methods that are designed for fixed camera, e.g. PAWCS, SuBSENSE, ViBe, cannot achieve accurate background subtraction on videos captured by moving camera. The deep learning model BSUV-Net, which can tackle unseen videos, also fails to detect the foreground in many videos of the customized dataset. The enhanced model BSUV-Net 2.0, which is trained with PTZ like augmented data, performs much better than the first version. SD-BMC, as compared with BSUV-Net 2.0, can detect more accurate saliency with fewer *FP* and *FN* errors.

## 5. Conclusion

We propose a new framework, SD-BMC, for the detection of salient regions in each video frames. The framework contains two major modules: video completion-based background modeler and the deep learning-based foreground segmenter network. In order to enable our framework for long-term saliency detection, the background modeler can adjust the background image dynamically via a feedback mechanism. SD-BMC can best segment foreground in videos captured by moving camera. To demonstrate this capability, we create our customized dataset with challenging videos captured by PTZ camera and handheld camera. The results, obtained from the PTZ videos, show that our proposed framework outperforms some deep learning-based background subtraction models by 11% or more. With more challenging videos, our framework also outperforms many high ranking background subtraction methods by more than 3%.

Although the results show that SD-BMC is superior to other deterministic as well as deep learning-based background subtraction methods, there are still ways to further improve it. In this work, we focus on designing the saliency detection framework for moving camera videos. In the future, we will work on new models for other challenging scenarios. The background modeling process can be made faster in order to tackle abrupt changes. Also, the foreground segmenter can adopt the teacher-student structure. While the complex teacher model is used in the training process, the testing process will be performed by a simpler student model. The lite model can be used for real-time saliency detection.

**Acknowledgment**
This work is supported by Hong Kong Innovation and Technology Commission and City University of Hong Kong (Project No. 9042823).